\documentclass[letterpaper, 10 pt, journal, twoside ]{IEEEtran}
\IEEEoverridecommandlockouts
\usepackage[utf8]{inputenc}
\usepackage[english]{babel}
\usepackage[T1]{fontenc}
\usepackage{amssymb,amsfonts}
\usepackage[cmex10]{amsmath}
\usepackage{dsfont}

\usepackage[noend]{algpseudocode}
\usepackage[ruled,vlined,linesnumbered]{algorithm2e} 
\SetKwInOut{Data}{Data}
\usepackage{multirow}
\usepackage{graphicx}
\usepackage{siunitx}
\usepackage{tikz}
\usepackage[
   colorlinks,
   linkcolor={red!50!black},
   citecolor={blue!50!black},
   urlcolor={blue!80!black}
]{hyperref}
\usepackage[caption=false,font=footnotesize]{subfig}

\usepackage[capitalize]{cleveref}
\crefformat{equation}{(#2#1#3)}
\crefrangeformat{equation}{(#3#1#4) to~(#5#2#6)}
\crefmultiformat{equation}{(#2#1#3)}%
{ and~(#2#1#3)}{, (#2#1#3)}{ and~(#2#1#3)}

\def\BibTeX{{\rm B\kern-.05em{\sc i\kern-.025em b}\kern-.08em
    T\kern-.1667em\lower.7ex\hbox{E}\kern-.125emX}}

\usepackage[normalem]{ulem}  

\newcommand{\tmin}{\ensuremath{\text{t}\textsubscript{min}}}
\newcommand{\tmax}{\ensuremath{\text{t}\textsubscript{max}}}
\newcommand{\trep}{\ensuremath{\text{t}\textsubscript{replan}}}
\newcommand{\Tmax}{\ensuremath{\text{C}\textsubscript{max}}}

\newcommand{\Vmax}{\ensuremath{\text{v}\textsubscript{max}}}
\newcommand{\Amax}{\ensuremath{\text{a}\textsubscript{max}}}
\newcommand{\Jmax}{\ensuremath{\text{j}\textsubscript{max}}}

\newcommand{\matr}[1]{\ensuremath{\mathbf{#1}}}

\newcommand{\boundellipse}[3]
{(#1) ellipse (#2 and #3)
}

\newcommand{\disable}[1]{}
\newcommand\norm[1]{\lVert#1\rVert}

\usepackage{scalerel}
\usetikzlibrary{svg.path}
\definecolor{orcidlogocol}{HTML}{A6CE39}
\tikzset{
orcidlogo/.pic={
  \fill[orcidlogocol] svg{M256,128c0,70.7-57.3,128-128,128C57.3,256,0,198.7,0,128C0,57.3,57.3,0,128,0C198.7,0,256,57.3,256,128z};
  \fill[white] svg{M86.3,186.2H70.9V79.1h15.4v48.4V186.2z}
               svg{M108.9,79.1h41.6c39.6,0,57,28.3,57,53.6c0,27.5-21.5,53.6-56.8,53.6h-41.8V79.1z M124.3,172.4h24.5c34.9,0,42.9-26.5,42.9-39.7c0-21.5-13.7-39.7-43.7-39.7h-23.7V172.4z}
               svg{M88.7,56.8c0,5.5-4.5,10.1-10.1,10.1c-5.6,0-10.1-4.6-10.1-10.1c0-5.6,4.5-10.1,10.1-10.1C84.2,46.7,88.7,51.3,88.7,56.8z};
}
}
\newcommand\orcidicon[1]{\href{https://orcid.org/#1}{\mbox{\scalerel*{
\begin{tikzpicture}[yscale=-1,transform shape]
\pic{orcidlogo};
\end{tikzpicture}
}{|}}}}

\newcommand{\PREPRINTYEAR}{2024}
\newcommand{\PUBLISHEDIN}{IEEE Robotics and Automation Letters}
\newcommand{\DOI}{10.1109/LRA.2024.3518096} 

\usepackage[placement=top,vshift=-7]{background}
\SetBgScale{1.0}
\SetBgContents{\PUBLISHEDIN. PREPRINT VERSION - DO NOT DISTRIBUTE. \href{https://doi.org/\DOI}{DOI \DOI}}
\SetBgColor{black}
\SetBgAngle{0}
\SetBgOpacity{1.0}

\begin{document}

\thispagestyle{empty}
\onecolumn
{
  \topskip0pt
  \vspace*{\fill}
  \centering
  \LARGE{%
    \copyright{} \PREPRINTYEAR~\PUBLISHEDIN\\\vspace{1cm}
    Personal use of this material is permitted.
    Permission from \PUBLISHEDIN~must be obtained for all other uses, in any current or future media, including reprinting or republishing this material for advertising or promotional purposes, creating new collective works, for resale or redistribution to servers or lists, or reuse of any copyrighted component of this work in other works.}
    \vspace*{\fill}
}
\NoBgThispage
\twocolumn          	
\BgThispage

\markboth{IEEE Robotics and Automation Letters. Preprint Version. Accepted November, 2024}
{Ghotavadekar \MakeLowercase{\textit{et al.}}: Variable Time-step MPC for Agile Multi-rotor UAV Interception of Dynamic Targets} 

\title{Variable Time-step MPC for Agile Multi-rotor UAV Interception of Dynamic Targets}

\author{Atharva Ghotavadekar$^{1}$$^{\orcidicon{0009-0000-8490-8988}}$,  \and František Nekovář$^{2}$$^{\orcidicon{0000-0002-1975-078X}}$, \and Martin Saska$^{2}$$^{\orcidicon{0000-0001-7106-3816}}$, \and  Jan Faigl$^{2}$$^{\orcidicon{0000-0002-6193-0792}}$
\thanks{Manuscript received: June, 07, 2024; Revised November, 09, 2024; Accepted November, 27, 2024.} 
\thanks{This paper was recommended for publication by Editor Ashis Banerjee upon evaluation of the Associate Editor and Reviewers' comments.
This work was supported by the Czech Science Foundation (GAČR) under research projects No. 22-05762S and No. 22-24425S, the CTU grant No. SGS23/177/OHK3/3T/13 and by the European Union under the project Robotics and advanced industrial production No. CZ.02.01.01/00/22\_008/0004590.}
\thanks{$^{1}$Author is with the Department of Electrical and Electronics Engineering, BITS Pilani K.K. Birla Goa Campus, India {\tt\footnotesize f20201300@goa.bits-pilani.ac.in}}%
\thanks{$^{2}$Authors are with the Czech Technical University, Faculty of Electrical Engineering, Technicka 2, 166 27, Prague, Czech Republic, {\tt\footnotesize \{nekovfra|saskam1|faiglj\}@fel.cvut.cz}}
\thanks{Digital Object Identifier (DOI): see top of this page.}
}

\maketitle
\begin{abstract}
   Agile trajectory planning can improve the efficiency of multi-rotor Uncrewed Aerial Vehicles (UAVs) in scenarios with combined task-oriented and kinematic trajectory planning, such as monitoring spatio-temporal phenomena or intercepting dynamic targets.
Agile planning using existing non-linear model predictive control methods is limited by the number of planning steps as it becomes increasingly computationally demanding.
That reduces the prediction horizon length, leading to a decrease in solution quality.
Besides, the fixed time-step length limits the utilization of the available UAV dynamics in the target neighborhood.
In this paper, we propose to address these limitations by introducing variable time steps and coupling them with the prediction horizon length.
A simplified point-mass motion primitive is used to leverage the differential flatness of quadrotor dynamics and the generation of feasible trajectories in the flat output space.
Based on the presented evaluation results and experimentally validated deployment, the proposed method increases the solution quality by enabling planning for long flight segments but allowing tightly sampled maneuvering.

\end{abstract}

\begin{IEEEkeywords} 
Aerial Systems: Applications, Motion and Path Planning, Autonomous Vehicle Navigation
\end{IEEEkeywords}

\section{Introduction}
\IEEEPARstart{T}{he} task of multi-target visit, monitoring, and interception planning under a flight budget constraint can be modeled as a variant of the \emph{Orienteering Problem} (OP)~\cite{op}.
The OP is to maximize the collected reward by visiting a known set of targets with the associated rewards within the given limited travel budget; thus, it can be thought of as a combination of two NP-hard combinatorial problems, the \emph{Travelling Salesman Problem} (TSP) and Knapsack problem~\cite{pisinger1998}.
Solutions to the OP generalizations~\cite{op_variants}, including the Team OP \cite{chao1996team}, Set OP \cite{set_op}, and Dubins OP\cite{dubins_op} can be effectively transferred to real-world use-cases by addition of constraints adhering to the specific vehicle dynamics.
In particular, the \emph{Kinematic OP} (KOP)~\cite{kop, meyer2023top} with limited velocity and acceleration-bound point-mass is especially suited for the dynamics of the multi-rotor \emph{Uncrewed Aerial Vehicles}~(UAVs).
Furthermore, a spatio-temporal reward-collection model is suitable for addressing the real-world dynamics of the targets to be visited and thus improving vehicle utilization under flight time constraints.

\begin{figure}[t]\centering
    \includegraphics[width=0.86\columnwidth]{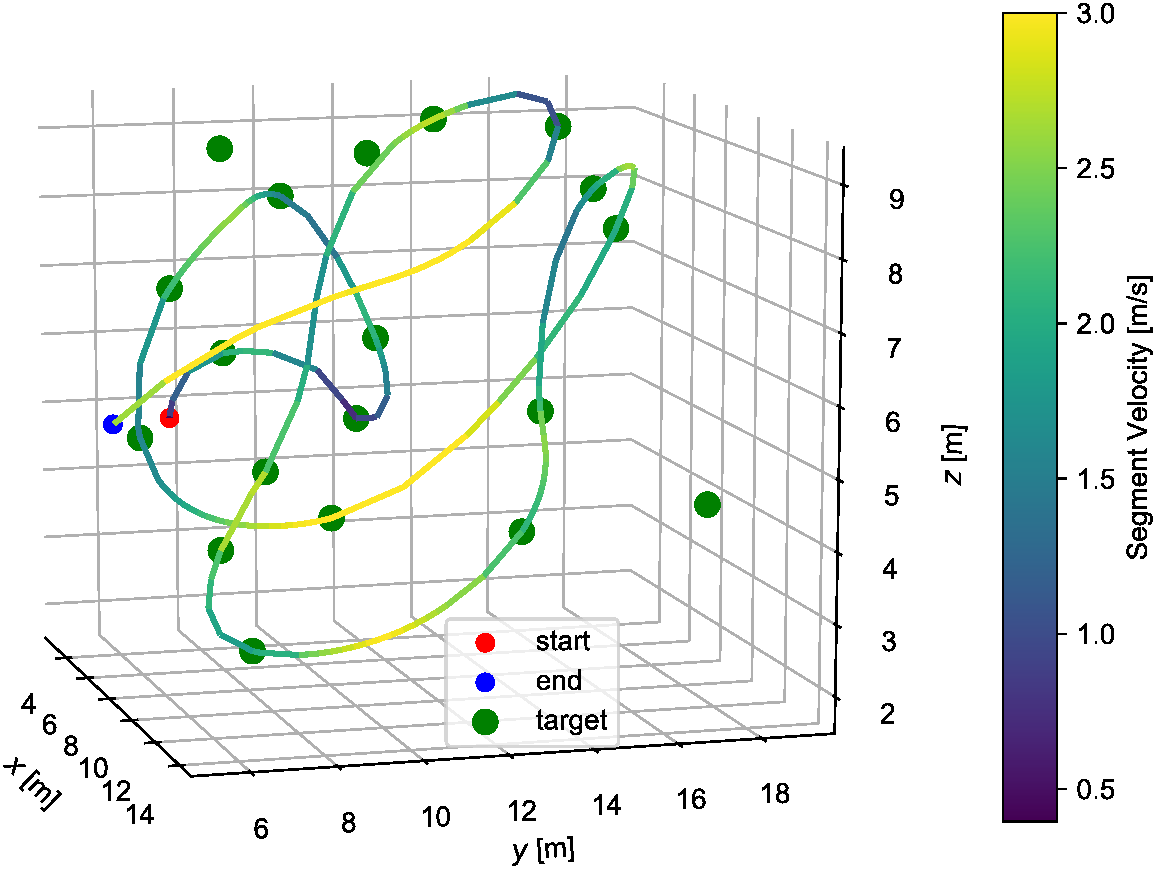}
    \caption{A solution to a 3D variant of the Tsiligirides 2 benchmark instance of the KOP determined by the proposed variable time-step method with the travel budget of \SI{40}{\second} with the visualization of the vehicle velocity for each prediction step.}
    \label{fig:cover-page}
    \vspace{-1.2em}
\end{figure}

The herein studied problem is motivated by dynamic environment monitoring formulated as a 3D generalization of the KOP using \emph{Model Predictive Control} (MPC)~\cite{impdr}.
The environment is defined by a set of target reward locations placed in a 3D environment representation, each location with the associated reward gain value collected when the location is visited.
However, motivated by data gathering tasks~\cite{monitoring_survey}, such as monitoring forest fires~\cite{forest_fire_monitoring}, coastal regions~\cite{coastal_dune_monitoring}, terrain surveillance~\cite{terrain_monitoring}, and urban inspections~\cite{road_tracking}, we are accounting for locations and their associated information gained that are highly dynamic due to changing environmental conditions.
Hence, the targets' dynamics is modeled in the proposed solution of the addressed variant of the KOP.
Since the target dynamics estimation is considered beyond the scope of the paper, it is assumed to be known, provided by an external system or an independent estimator.

The necessity to account for target dynamics puts high demands on trajectory planning, where non-linear MPC-based methods suffer from the fixed length time-step, which significantly increases the computational burden for an increased planning horizon.
Since these methods can exploit the vehicle dynamics, we propose a variable time-step approach to increase the length of the prediction horizon without increasing the number of prediction steps.
Based on the evaluation results for benchmark instances of the KOP, the proposed method provides improved solutions compared to the existing methods.
Furthermore, the practical applicability of the method has been experimentally validated in real-field deployments.

The rest of the paper is organized as follows.
An overview of the related methods is provided in the following section.
The problem is formally introduced in \cref{sec:problem} together with the models of the target and reward dynamics, vehicle model, and the utilized cost function.
The proposed online replanning is presented in \cref{sec:method}.

\section{Related Work}

The studied problem combines trajectory planning with routing, where a multi-point trajectory needs to be determined for a particular sequence of visits to the target regions.
Since such a trajectory is sequence-dependent, agile trajectory planning depends on trajectory parameterization, allowing the exploitation of vehicle dynamics.
Pioneered quadrotor trajectory generation methods are minimum snap polynomial approaches based on the differential flatness of quadrotor dynamics~\cite{minsnap}.
B-spline~\cite{bspline_planning, bspline_replanning} and Bézier curve~\cite{bezier_faigl, bezier_close_enough_op} based approaches produce spatial polynomial paths that require mapping to time domain~\cite{optimal_time_allocation_icra, pa_path_parameterization_icra2020}. 
In~\cite{optimal_time_allocation_icra}, it is shown that if the initial and final velocities are non-zero, a feasible solution may not exist with the aforementioned approaches.

On the other hand, routing problems, as a variant of the TSP~\cite{survey_routing}, are required to adhere to real-world time and/or energy budget constraints and are modeled as the OP~\cite{op}.
Dubins vehicle is a simplified motion model with constant velocity constrained by a minimum turning radius and a maximum radial acceleration already deployed in Dubins routing problems~\cite{dubins_op,close_enough_dop}.
However, the advantage of energy efficiency in Dubins paths is offset by possible significant detours, and the model's restriction is more relevant for fixed-wing UAVs than for agile multi-rotor UAVs.

Time optimal trajectories based on Pontryagin's minimum principle that keeps control input at their limits in a bang-zero-bang profile can be calculated axis-wise and then synchronized~\cite{Romero_2022}.
The authors of~\cite{top_pmm_failure_behnke} found that an analytical solution for axis synchronization does not exist for certain instances for a jerk-input point-mass vehicle model.
In~\cite{kop,meyer2023top}, it is further shown that even for the acceleration-input point-mass model, axis synchronization sometimes results in failures due to improper consideration of the robots' inertia, and a method to overcome the synchronization failure is proposed.
However, the control input is limited by axis-wise kinematic constraints, the aggregation of which results in a $n$-dimensional cuboid.
In~\cite{segmented_tmpc}, the authors optimize for segment-wise sampling time for a pre-assigned fixed number of time steps and trajectories passing through fixed waypoints.
However, the approach does not address the selection of waypoints for maximum reward collection or moving-horizon planning.

\begin{figure}[t]\centering
   \includegraphics[width=0.98\columnwidth]{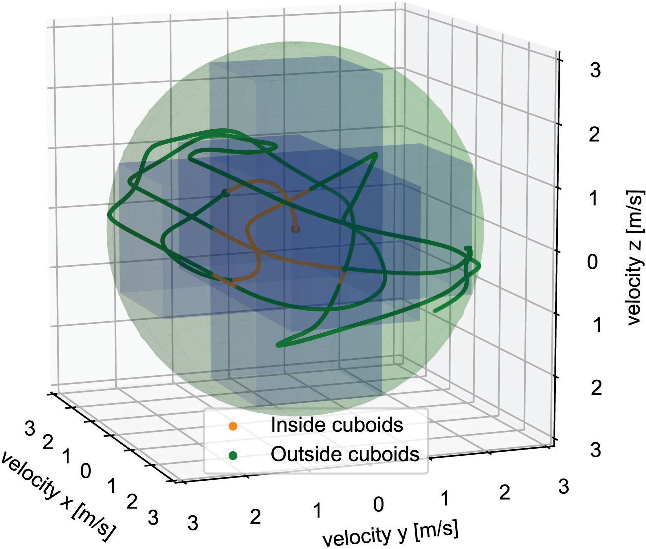}
   \caption{Analysis of the velocity profile from the trajectory depicted in~\cref{fig:cover-page} with respect to (w.r.t.) the kinematic constraints imposed by \cite{meyer2023top} (blue cuboids) and constraints imposed by the proposed method (green sphere). 
   About \SI{84.4}{\percent} of the velocity profile (green trajectory) lies outside the cuboidal constraints but within the proposed spherical constraints.}
   \label{fig:kinematic_constraints}
\end{figure}
The proposed approach overcomes the aforementioned limitations to maximize the reward collection and utilization of the multi-rotor agility in static routing problems and continual monitoring tasks.
We propose to utilize the full range of feasible kinematics, equivalent to an $n$-dimensional sphere with the radius $v_{max}$ and $a_{max}$, while enforcing velocity and acceleration constraints, respectively. 
It enables maximum utilization of multi-rotor kinematic constraints as shown in~\cref{fig:kinematic_constraints}.
The contributions are considered as follows.
\begin{itemize}
   \item 3D KOP generalization with a viable jerk-input constrained trajectory planner for multi-rotor UAV platforms.
   \item A novel optimal control formulation called the \emph{Variable Time-step MPC} (VT-MPC), where the prediction time-step is a part of the variables optimized over the control horizon in reward-collection planning, which improves performance in the KOP and related receding-horizon monitoring problems.
   \item The VT-MPC formulation-based planning and executing optimal reward-collecting trajectories for monitoring tasks in real-world flight that leverages the differential flatness property of the quad-copter dynamics combined with a~simplified motion primitive.
   \item Comparative evaluation of the proposed method and state-of-the-art methods to the KOP showing improved performance along with the online re-planning ability of the proposed formulation in a receding-horizon planning.
   \item Implementation of the method as open source code supporting replicability of the reported results.
\end{itemize}


\section{Proposed Models and Problem Formulation}\label{sec:problem}

The studied problem and the proposed novel VT-MPC formulation are inspired by previous work on water surface monitoring~\cite{impdr}.
Therefore, we follow therein introduced notation further extended to address the VT-MPC.
\begin{itemize}
   \item $t_k$ is the temporal length of the $k$th prediction step.
   \item $N_t$ is the number of targets in the 3D environment.
   \item $N_h$ is the number of the prediction steps.
   \item $\mathbf{p}_k$ is the position vector denoting the location of the UAV at the $k$th step.
   \item $\mathbf{q}_{r,k}$ is the position vector of the $r$th target at the $k$th step.
   \item $g_{r,k}$ is the reward gain value associated with the $r$th target at the $k$th step,
   \item $\mathbf{x}_k$ denotes the state of the UAV at the $k$th step.
   \item $\mathbf{u}_k$ is the input of the model at the $k$th step.
\end{itemize}
In the rest of the section, the formal models are introduced for target and reward dynamics (\cref{sec:problem_target}), vehicle modeling in \cref{sec:problem_vehicle}, and cost function in \cref{sec:problem_cost}.
The proposed problem formulation is presented in \cref{sec:problem_optimal}.

\subsection{Target and Reward Dynamics}\label{sec:problem_target}
The target locations with the associated reward gains at the prediction step $k$ are denoted by the vectors $\mathbf{q}_k$ and $\mathbf{g}_k$, respectively.
For the $r$th target, $\mathbf{q}_{r,k}=[q_{r,k}^x, q_{r,k}^y, q_{r,k}^z]$ represents the target's position vector, and $\mathbf{q}_k=[\mathbf{q}_{1,k}, \ldots, \mathbf{q}_{N_t,k}]$ represents combined target position vector.
Thus, the Euclidean distance of the UAV at the position $\mathbf{p}_k$ from the target $\mathbf{q}_r$ at the $k$th prediction step can be computed as  
\begin{equation} \label{eq:uav_distance}
d_{r,k}(\mathbf{p}_k, \mathbf{q}_{r,k}) = \sqrt{(p_{k}^x - q_{r,k}^x)^2 + (p_{k}^y - q_{r,k}^y)^2 + (p_{k}^z - q_{r,k}^z)^2}.
\end{equation} 
Assuming that target position dynamics are known, we denote the target position vector at the prediction step $k$ as $\mathbf{q}_k$.

The information collecting sensor is modeled using the differentiable Butterworth function denoted by $f_b$, with the order $n_b$ affecting the function shape and cutoff parameter $c_b$ that models the approximate sensor collection range as
\begin{equation} \label{eq:sensor_function}
f_b(x) = \frac{1}{1 + \bigl(\frac{x}{c_b}\bigl)^{n_b}}.
\end{equation}

The dynamics of the reward gain dynamic state $g_{r,k}$ associated with the $r$th target at the $k$th step is determined as
\begin{equation}\label{eq:reward_dynamics}
   g_{r,k+1} = \Bigl(g_{r,k} + \alpha_g t_k \Bigl) \cdot \Bigl(1 - f_b(d_{r,k})\Bigl),
\end{equation}
where $\alpha_g$ is the rate of linear increase of the reward gain in time.
On the UAV visit of the $r$th target at the $k$th step, the distance value $d_{r,k} = 0$ causes the term $1-f_b(d_{r,k})$ to be nullified; hence, the reward is assumed as fully collected.
The used model of information dynamics (to be collected within neighborhoods of discrete points') allows formulating the multiple-target monitoring task as an MPC, with a cost function minimizing the reward gain values of all the targets with regard to the modeled vehicle dynamics. The reward gain is modeled to increase with time when uncollected, to facilitate formulation of persistent monitoring tasks.

\subsection{Vehicle Model}\label{sec:problem_vehicle}

The utilized vehicle model is based on a point-mass motion primitive that is able to produce dynamically feasible UAV trajectories.
States and inputs to our discretized algebraic system are considered as the differentially flat output states of multi-rotor dynamics~\cite{minsnap}.
The UAV state $\mathbf{x_k}$ at the $k$th prediction step is denoted 
\begin{equation}\label{eq:uav_state_vec}
   \mathbf{x}_{k} = [\mathbf{p}_k, \mathbf{v}_k, \mathbf{a}_k, \psi_k],
\end{equation}
where the respective elements are position, velocity, acceleration vectors, and heading.
We introduce the variable time-step $t_k$ as a system input
\begin{equation}\label{eq:motion_primitives}
   \mathbf{u}_k = [\mathbf{j}_k, \dot\psi_k, t_k],
\end{equation}
where the respective input elements are jerk, heading rate, and the time-step length optimization variable, which is also referred to as the variable time-step.

\begin{figure}[t]\centering
   \includegraphics[width=0.46\textwidth]{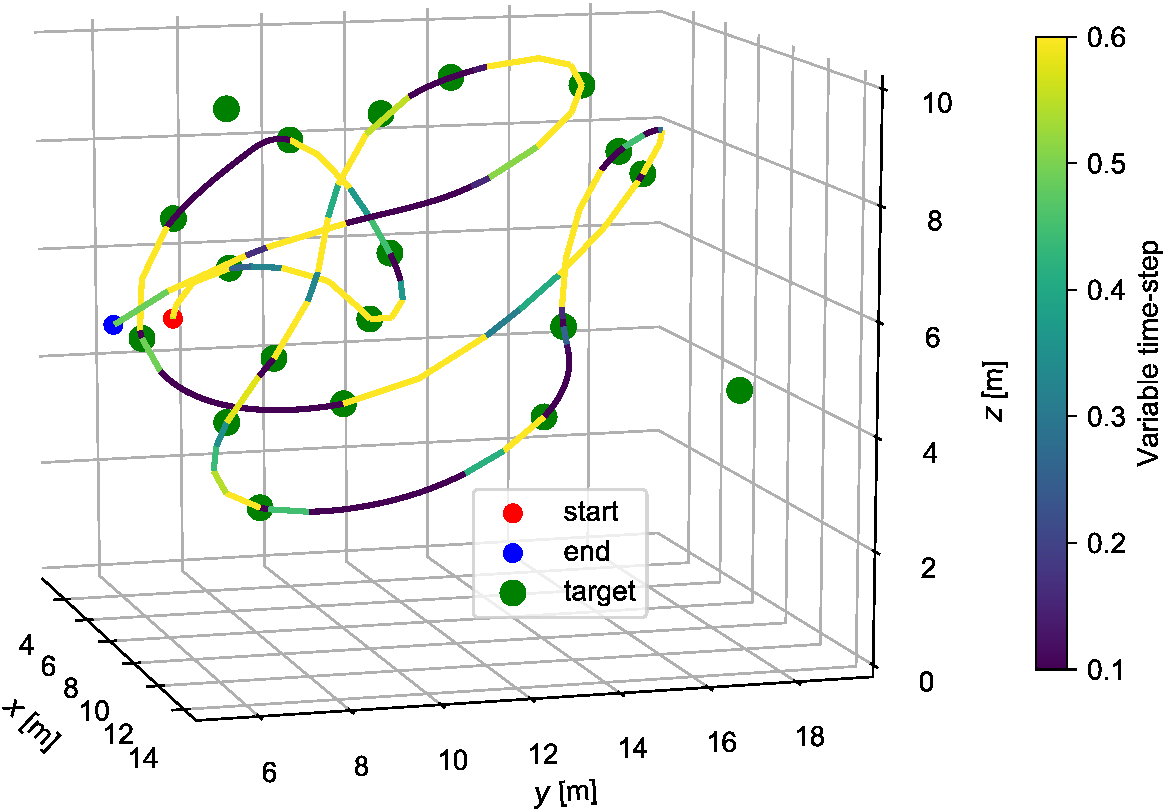}
   \caption{%
      Example of the variable time-step lengths distribution for a data collecting trajectory of the Tsiligirides 2 instance extended to 3D.
      The optimal solution utilizes shorter time-steps in areas of reward collection and frequent input changes due to maneuvering.}
      \label{fig:tsili_2_3d_timestep_profile}
\end{figure}
An example of the variable time step for a vehicle trajectory is depicted in \cref{fig:tsili_2_3d_timestep_profile}.
The discretized point-mass model with the jerk input is used for state propagation according to \cref{eq:state_propagation}.
\begin{equation}
   \begin{array}{lcl}\label{eq:state_propagation}
      \mathbf{a}_{k+1} &= &\mathbf{a}_{k}+ \mathbf{j}_k\cdot t_k  \\
      \mathbf{v}_{k+1} &= &\mathbf{v}_{k}+ \mathbf{a}_k \cdot t_k + \frac{1}{2}\cdot \mathbf{j}_k \cdot t_k^2 \\
      \mathbf{p}_{k+1} &= &\mathbf{p}_{k}+ \mathbf{v}_k \cdot t_k + \frac{1}{2}\cdot \mathbf{a}_k \cdot t_k^2 + \frac{1}{6} \cdot \mathbf{j}_k\cdot t_k^3\\
      {\psi}_{k+1} &= &{\psi}_{k} + {\dot\psi}_{k} \cdot {t_{k}}
   \end{array}
\end{equation}

\subsection{Cost Functions}\label{sec:problem_cost}
The sum of the reward gains for all targets is directly a~part of the MPC cost function.
Besides, the heading cost $c_\psi(\psi_k)$ is included in the cost function since the multi-rotor vehicle allows for independent heading control, which can be added to the perception constraints to meet the mission objectives.
Thus, $c_\psi(\psi_k)$ is defined as the negative of the dot product of the UAV velocity with the current heading in the X-Y plane to ensure the UAV faces its direction of motion.
The change in the input is penalized by a scalar $c_u$.
The cost function $c_{k}$ at the prediction step $k$ is thus defined as
\begin{equation}\label{eq:mpc_cost_ftion}
   c_{k} =  \displaystyle \sum_{r=0}^{N_t} (g^r_k) + k_\psi \cdot c_\psi(\psi_k) + \Delta \mathbf{u}^T_k c_u \Delta \mathbf{u}_k,
\end{equation}
where $k_\psi$ is a tunable scaling parameter.
The summation term denotes the total uncollected reward at the $k$th step.

A~soft constraint on the final position is imposed via the terminal cost $c_f$ defined as
\begin{equation}\label{eq:mpc_terminal_ftion}
   c_{f} =  \displaystyle k_f \cdot d(\mathbf{p}_{N_h}, \mathbf{p}_{f}),
\end{equation}
where $\mathbf{p}_f$ is the desired final UAV position.

\subsection{Optimal Control Problem}\label{sec:problem_optimal}

The combined state vector of the system $\mathbf{X}_k$ is defined in~\cref{eq:state_vector} as the concatenated vector of the reward positions $\mathbf{q_k}$, reward gains $\mathbf{g_k}$, the UAV state vector $\mathbf{x_k}$, and the scalar $t\textsubscript{sum,$k$} = \sum_{i=0}^k t_i$ denoting the total trajectory time (temporal length) up to the prediction step $k$.
\begin{equation}\label{eq:state_vector}
  \mathbf{X}_k = [\mathbf{q}_k, \mathbf{g}_k, \mathbf{x}_k, t\textsubscript{sum,$k$}]
\end{equation}

The optimal control problem is formulated as the VT-MPC with the prediction horizon of $N_h$ steps
\begin{equation}\label{eq:optimal_control_problem}
   \begin{array}{rrclcl}
      \displaystyle \min_{\mathbf{X}, \mathbf{u}} & \multicolumn{3}{l}{\displaystyle \sum_{k=0}^{N_h}(c_{k}(\mathbf{X}_k,\mathbf{u}_k)) +  c_{f}} \\
      \textrm{s.t.} & \norm{\mathbf{v}_k} \leq \Vmax \\
      & \norm{\mathbf{a}_k} \leq \Amax \\
      & -\Jmax \leq \mathbf{j}_k \leq \Jmax \\
      & -\dot\psi_{\text{max}} \leq \dot\psi \leq \dot\psi_{\text{max}} \\
      & \tmin \leq t_k \leq \tmax \\
      & t\textsubscript{sum,$N_h$} = \Tmax\\
   \end{array}.
\end{equation}
The UAV platform's specific kinematic bounds are $v_{\text{max}}$, $a_{\text{max}}$, $j_{\text{max}}$, and $\dot\psi_{\text{max}}$ on the velocity, acceleration, acceleration jerk, and heading rate, respectively.
Time-step length bounds are $t_{\text{min}}$ and $t_{\text{max}}$, and the final equation constrains the travel budget \Tmax.

\begin{figure}[!ht]\centering\vspace{-1em}
\includegraphics[width=0.5\textwidth]{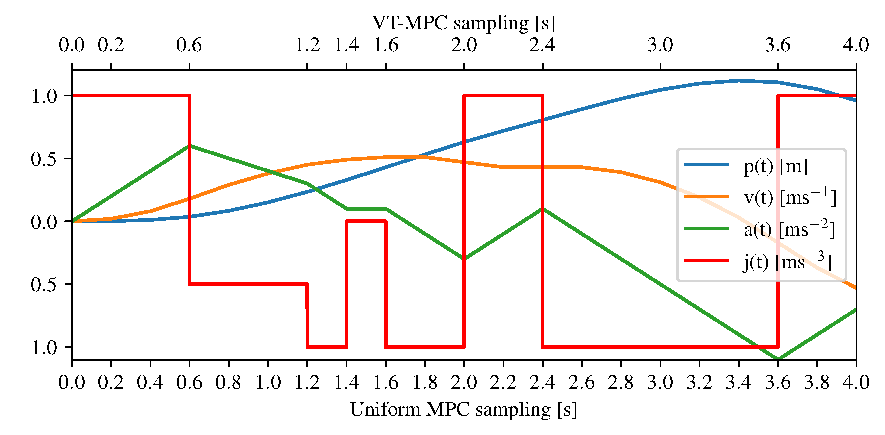}
\caption{%
  An example single-dimensional jerk-input trajectory where time-step sampling of the VT-MPC is shown as the top x-axis and of the MPC as the bottom x-axis.
  With the MPC times-step of $\SI{0.2}{s}$, the formulation requires \num{20} prediction steps to reach full fidelity on the prediction horizon $\SI{4}{s}$.
  The VT-MPC set at $t\textsubscript{min}=\SI{0.2}{s}$, $t\textsubscript{max}=\SI{0.6}{s}$ and $t\textsubscript{sum}=\SI{4}{s}$ utilizes only \num{10} prediction steps. The first optimization time-step is always constrained at $t\textsubscript{min}$.
\label{fig:traj_sampling}}\vspace{-1em}
\end{figure}
The difference between standard MPC and the proposed VT-MPC planning method is illustrated in~\cref{fig:traj_sampling}.
The MPC models a discrete dynamical system where inputs to the system can change only at fixed time-steps.
Assuming a mass-point vehicle model where all dynamical states are constrained, we note the optimization can converge to a situation where constraints on the input or some dynamical state are reached, leading to constant or zero input value respectively on multiple subsequent times-steps.
It is supported by the observation that a time-optimal position-velocity trajectory neglecting higher-order dynamics would intercept targets using straight lines, and is a lower-bound to our more complex trajectory with gap decreasing with higher target distances and/or tighter constraints.
We take advantage of the observation by formulating the variable time-step, where limited number of prediction steps are optimized on the same time horizon.
\section{Proposed Online Re-planning}\label{sec:method}

The proposed problem formulation can be used for offline trajectory planning and online re-planning for real-world tasks in the presence of target dynamics, including reward gain dynamics.
However, the prediction horizon $N_h$ and the travel budget \Tmax{} need to be adjusted to fit the receding horizon fashion for continual monitoring tasks.
Besides, a new parameter is introduced to define the re-planning period \trep{} to define the maximum time used for the onboard re-planning.

The targets to be monitored exhibit known spatio-temporal dynamics that the planner uses for precise interception.
In the absence of the target state measurement, a simulator runs onboard to provide the reward state estimates (the target locations and their reward gains).
The optimizer is bound by a maximum allowed planning time equal to \SI{85}{\percent} of \trep{} to converge.
The reserve in the amount of \SI{15}{\percent} enables other computations and control routines within the onboard system.
Once a trajectory is planned, it is sent to the trajectory tracker~\cite{mpc_tracker} of the employed MRS UAV System~\cite{mrs_uav_sys}.


\begin{algorithm}[!htb]
   \caption{Proposed Online re-planning.}\label{alg:proposed}
   \DontPrintSemicolon
   \KwIn{
      \trep -- re-planning time in seconds,
      \matr{x}\textsubscript{0} -- initial model state,
      $n_{\text{reuses}}$ -- maximum number of trajectory reuses.
      }
      \Data{
	 \texttt{c} -- boolean indicating if the solution converged,
  \texttt{c}\textsubscript{use} -- trajectory reuse counter,
	 $\mathbf{x}$ -- the UAV state vector \cref{eq:uav_state_vec},
	 $S_{\text{p}}$ -- planned trajectory.
      }
      \vspace{2pt}\hrule\vspace{2pt}
	    \texttt{c}, $S_{\text{p}} \gets \operatorname{optimize}(\matr{x}\textsubscript{0},\infty)$ \tcp*[f]{\scriptsize Initial planner call.}\;
      \texttt{c}\textsubscript{use} $ \gets 1$\; 
      \While{{\rm \texttt{c}\textsubscript{use} $ \leq n_{\text{reuses}}$}}{
	 \eIf{{\rm \texttt{not}} \texttt{c}}{   
  
   \hspace*{-1.2em} $\rhd$ \textit{Since the solution was not found in the previous iteration, continue tracking the previous trajectory for {\rm \trep{}} seconds.}\; 
	    \texttt{c}\textsubscript{use} $\gets$ \texttt{c}\textsubscript{use}$ + 1$ \tcp*[f]{\scriptsize Increment use counter.}\;
	 }{
	    \texttt{c}\textsubscript{use}$ \gets 1$ \tcp*[f]{\scriptsize Solution found, reset counter.}\;
	    $\operatorname{set\_tracker}(S_{\text{p}})$ \tcp*[f]{\scriptsize Track the trajectory}\;
	 }
   \hspace*{-1.2em} $\rhd$ \textit{Set the state for the next iteration.}\; 
	 $\matr{x} \gets S\textsubscript{p}[\texttt{c}_{\text{use}}\cdot\trep]$ \;
    \texttt{c}, $S_{\text{p}} \gets \operatorname{optimize}(\matr{x},\trep)$\label{line:optimize}\; 
      }
\end{algorithm}

The online re-planning is summarized in \cref{alg:proposed}.
It works in iterations, where a found trajectory is reused up to $n_{\text{reuses}}$ times if a new trajectory $S_\text{p}$ is not found in \trep{} (\cref{alg:proposed}, \cref{line:optimize}).
Once the tracker is set, it follows the trajectory unless a new trajectory is set.

The trajectory tracker produces a full state reference to be tracked by a low-level controller.
While the planned trajectory is being tracked by the UAV, the UAV state in \trep{} seconds in the future is used as the planner input for the next iteration.
That ensures continuity of the trajectories produced, despite the rate of re-planning being significantly lower than the rate of the onboard low-level control loop that is set to \SI{100}{\hertz}.
In the case a solver does not converge within the given planning time, the previously obtained trajectory is kept for the next period.

In the case the previously calculated trajectory is followed to its end,  the UAV stops and hovers at its current position to keep re-planning till a solution is found again.
Note that the future state reference and previously found solution can be used as an initial condition to warm-start the optimization by the \texttt{do\_mpc} framework~\cite{do_mpc} in the next iteration, thus reducing the computational burden.

The proposed online re-planning has been validated in simulation and real-world deployments, with the results reported in the following section.

\section{Results}

\begin{figure}[ht]\centering
   \includegraphics[width=0.95\linewidth]{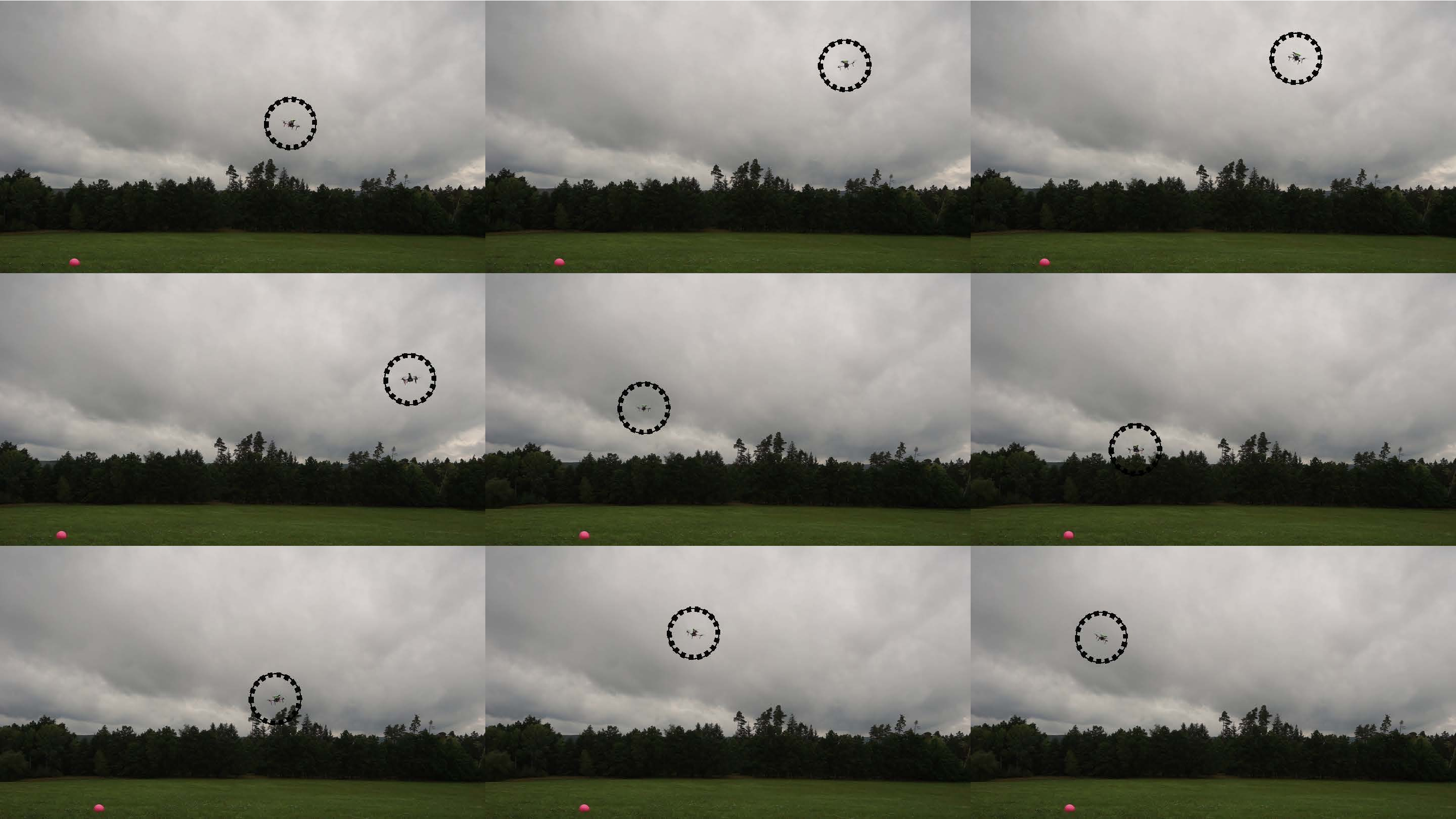}
   \caption[Hardware Experiment Snapshots]{Snapshots of the UAV performing the monitoring mission in a field test.
   The UAV is encircled in each image.}
   \label{fig:hardware_snapshots}
\end{figure}

The performance of the proposed approach has been evaluated in multiple simulations and real-world deployments to examine the impact of introducing the variable time step and to verify the dynamic feasibility of the trajectories generated.
2D and 3D KOP instance trajectories are computed offline on a consumer-grade desktop computer. 
The real-world feasibility and online re-planning ability is verified with multiple Gazebo simulations and on-field experiments using a UAV research platforms based on the T650, X500 and f450 UAV frames~\cite{hert2022mrs} using the MRS UAV System~\cite{mrs_uav_sys} running on the Intel NUC i7-8559U onboard computer, see real-world deployment snapshots in \cref{fig:hardware_snapshots}. 
The UAV is interfaced via \emph{Robot Operating System}~(ROS) and utilizes a standard GPS for localization.
All the computations are done onboard for real-time experiments involving re-planning.

The employed MPC framework~\cite{do_mpc} with IPOPT~\cite{ipopt} solver and the MA97 linear solver~\cite{hsl_collection} produces a system state prediction corresponding to the optimal control input over the prediction/control horizon.
The predicted UAV positions in the system state prediction are sent to the MRS UAV System interface for trajectory tracking after a temporal Cubic Spline interpolation. 
The system uses a linear MPC-based trajectory tracker~\cite{mpc_tracker} and a low-level controller that uses geometric state feedback in SE(3)~\cite{se3_controller} for precise reference tracking and execution of fast maneuvers.
The trajectories generated are verified to be dynamically feasible both in simulation and real-world experiments.

The reported results consist of simulation deployments in \cref{sec:results_simulation}, real-world deployments in persistent monitoring presented in \cref{sec:results_persistent}, and an assessment of the dynamic feasibility of the planned trajectories in~\cref{sec:results_dynamic}.

\subsection{Simulation Deployments}\label{sec:results_simulation}
Simulation results are presented for the 2D and 3D variants of the KOP for comparison of the proposed method with existing solutions to the KOP.
The proposed approach can tackle these variants by the removal of the target dynamics $(\alpha_g = 0)$ from~\cref{eq:reward_dynamics}, such that there is no change in the target reward gains unless collected by the monitoring UAV.
Additional terminal constraints on the UAV position and velocity are introduced to the problem, adapting the MPC formulation for a static single-run route planning task.
While the proposed method is employed for moving-horizon monitoring, the static KOP is suitable for comparison of performance metrics and the papers' main contribution (the variable timestep) can be better visualized in a static planning scenario.
The performance of the proposed VT-MPC is, therefore, evaluated against the existing static time-step MPC approach~\cite{impdr} and other available approaches to the mass-point KOP, which differ in kinematic constraint formulation.

The solution quality reflected as collected reward gains during UAV flight is compared with the state-of-the-art solutions represented by the KOP-1, KOP-6\textsuperscript{lns}~\cite{kop}, and MPC\textsubscript{static}~\cite{impdr}.
It is noteworthy that the original mass-point KOP solutions~\cite{kop} utilize identical kinematic constraints on per-axis velocity and acceleration, obtained by scaling magnitude constraints by a factor of $\sqrt{2}^{-1}$, due to inability to model vector magnitude constraints, and use accelerations as model inputs.
Kinematic constraints used by the MPC\textsubscript{static} and the VT-MPC formulations are $a_{\text{max}} = \SI{1.5}{m/s^2}$ on acceleration and $v_{\text{max}} = \SI{3.0}{m/s}$ on velocity vector magnitudes.
Additional input jerk per-axis constraints used for the VT-MPC are $j_{\text{max}} = \SI{30}{m/s^3}$, sensor function parameters are set to $c_b = 0.05$ and $n_b = 2$.

\begin{table}[h]\centering
\vspace{-1em}
   \caption{Collected reward gains for the 2D instances of the KOP.\label{tbl:resultsfirst}}\vspace{-1em}
%
\begin{tabular}{l r r r r r r r}
   \noalign{\hrule height 1.1pt}\noalign{\smallskip}
   $C_{\text{max}}$ &  $N_{\text{h}}$ &  KOP-1 & KOP-6\textsuperscript{LNS} & MPC\textsubscript{static} & \textbf{VT-MPC} & t\textsubscript{avg} \\
   \noalign{\smallskip}\hline\noalign{\smallskip}
   10 & 50 & 95 & 98 & 95 & \textbf{115} & 15.5\\
   15 & 75 & 180 & 180 & 200 & \textbf{215} & 27.1 \\
   20 & 67 & 250 & 250 & 230 & \textbf{345} & 38.2\\
   25 & 125 & 325 & 325 & 310 & \textbf{395} & 100.6\\
   30 & 150 & 390 & 390 & 380 & \textbf{430} & 120.8\\
   35 & 175 & 430 & 435 & \textbf{450} & \textbf{450} & 167.8\\
   40 & 200 & \textbf{450} & \textbf{450} & \textbf{450} & \textbf{450} & 238.6\\
   \noalign{\hrule height 1.1pt}
\end{tabular}

\end{table}
Results for the KOP instances of the 2D Tsiligirides 2 dataset~\cite{op} with the maximum available travel budget $C_\text{max} [s]$ are depicted in \cref{tbl:resultsfirst}.
For each budget, the instance was optimized 10 times to full solver convergence with reward values perturbed as in~\cite{impdr} with the average computational time reported as t\textsubscript{avg}.

{
   \begin{figure}[!htb]
      \subfloat[{$C_{\text{max}} = \SI{10}{\second}$}]{\includegraphics[width=0.49\columnwidth]{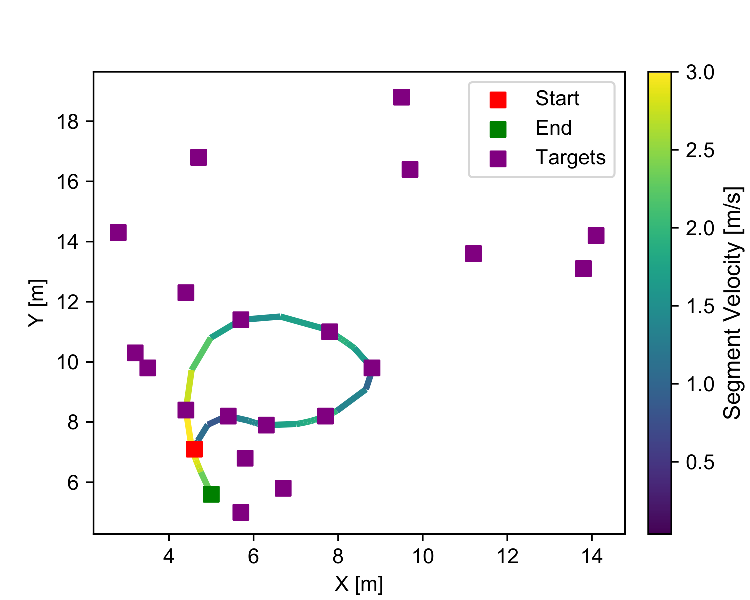}}
      \hfill
      \subfloat[$C_{\text{max}} = \SI{15}{\second}$]{\includegraphics[width=0.49\columnwidth]{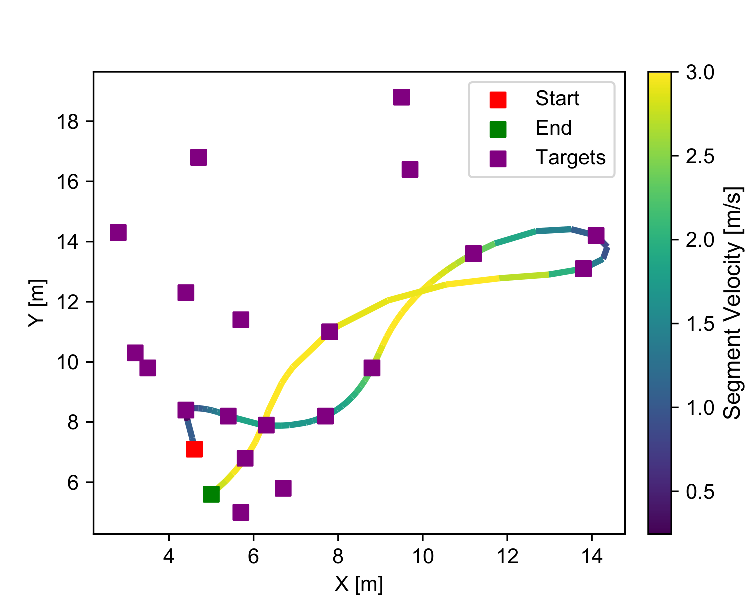}}

      \subfloat[$C_{\text{max}} = \SI{20}{\second}$]{\includegraphics[width=0.49\columnwidth]{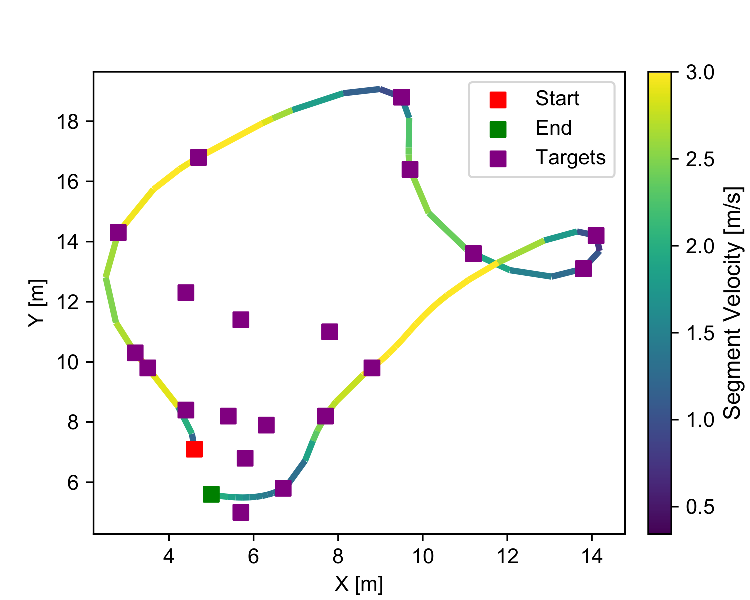}}
      \hfill
      \subfloat[$C_{\text{max}} = \SI{25}{\second}$]{\includegraphics[width=0.49\columnwidth]{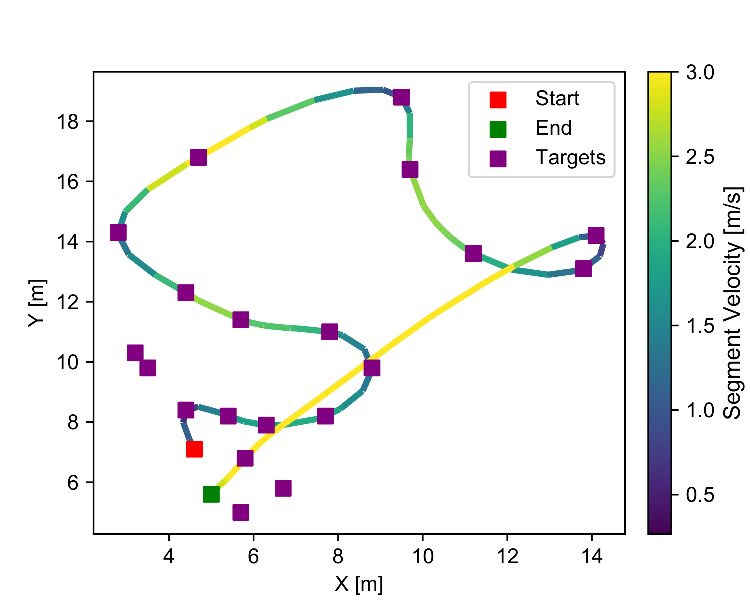}}
      \caption{Velocity profiles of the planned trajectories by the proposed VT-MPC planner on the instances from the Tsiligirides 2 dataset.}
      \label{fig:tsili_vel_profile}
   \end{figure}
   \begin{figure}[!htb]
      \subfloat[Predicted positions, $C_{\text{max}} = \SI{15}{\second}$]{\includegraphics[width=0.5\columnwidth]{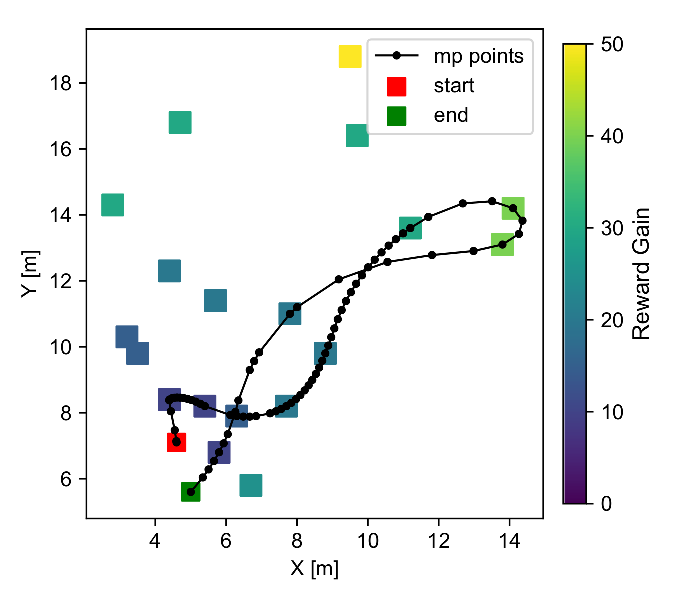}}
      \hfill
      \subfloat[Predicted positions, $C_{\text{max}} = \SI{25}{\second}$]{\includegraphics[width=0.5\columnwidth]{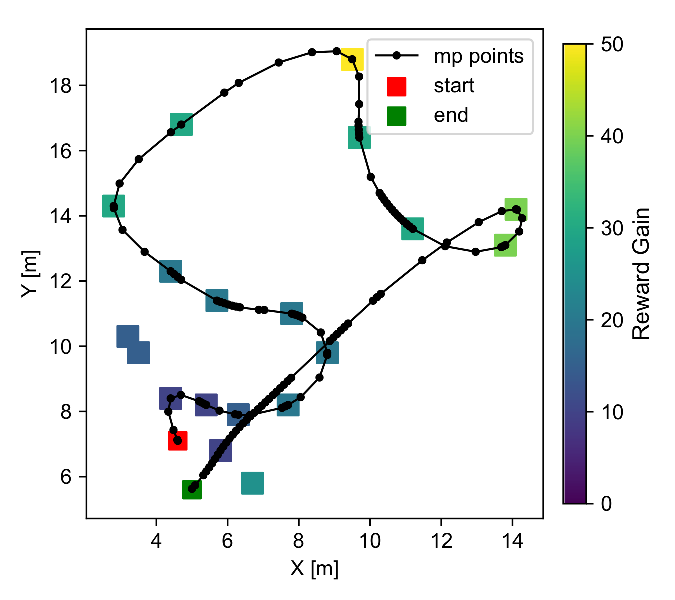}}

      \subfloat[Variable time-step, $C_{\text{max}} = \SI{15}{\second}$]{\includegraphics[width=0.5\columnwidth]{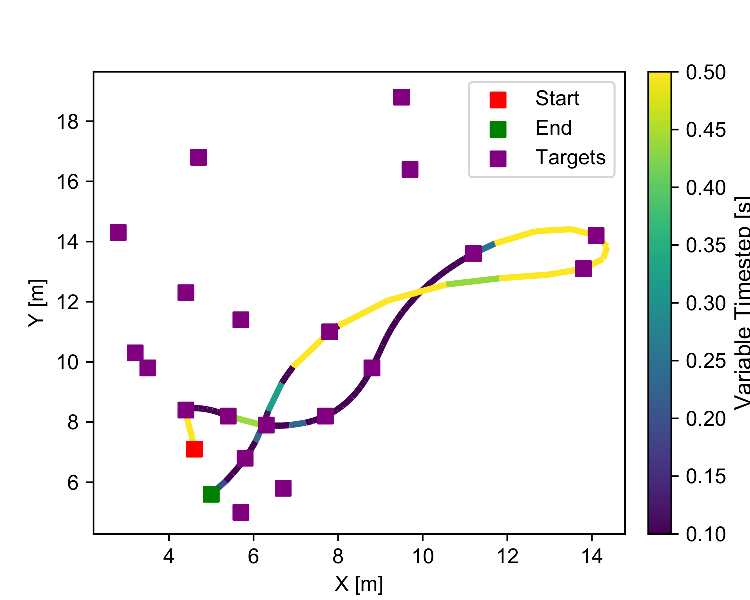}}
      \hfill
      \subfloat[Variable time-step, $C_{\text{max}} = \SI{25}{\second}$]{\includegraphics[width=0.5\columnwidth]{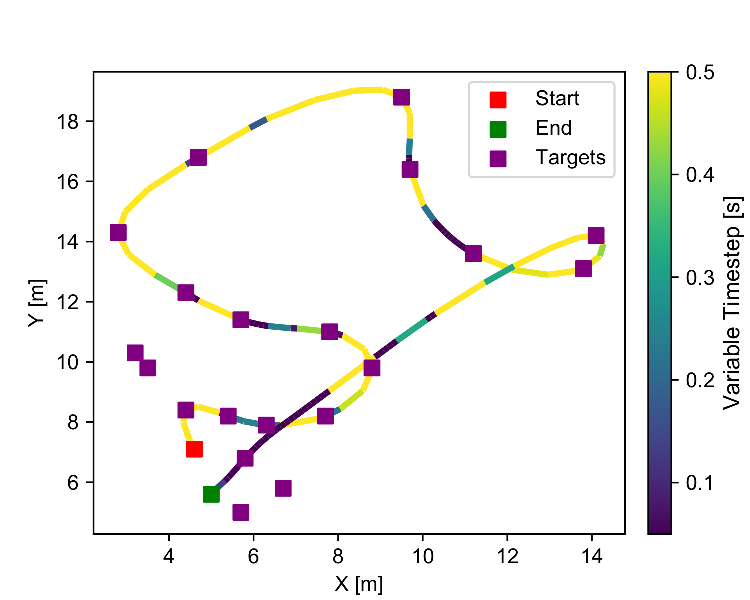}}

   \caption{The distribution of the predicted positions (upper row) and variable time-step (bottom row) on 2D trajectories of Tsiligirides 2 instances.}
   \label{fig:tsili_points_timestep_profile}
   \end{figure}
   }
   Selected velocity profiles of the dynamically feasible trajectories produced by the proposed planner are depicted in \cref{fig:tsili_vel_profile}.
   The distribution of sampled points and computed time-steps are shown in~\cref{fig:tsili_points_timestep_profile}.
   The same number of the prediction steps as in~\cite{impdr} was used for a fair comparison.




\begin{figure}[!htb]\centering
   \includegraphics[width=0.9\columnwidth]{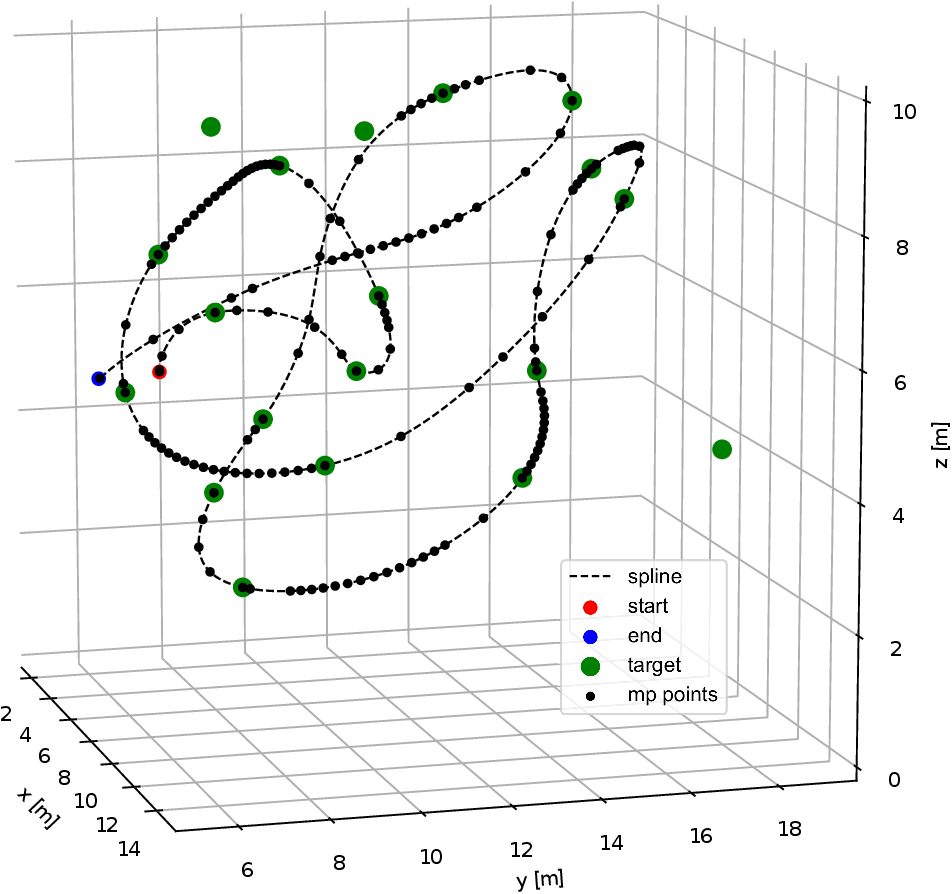}
   \caption{Distribution of predicted positions of the planned trajectory of the 3D Tsiligirides 2 instance with the travel budget $C_\text{max}=\SI{40}{\second}$.}
   \label{fig:tsili_3d}
   \vspace{-1em}
\end{figure}
Finally, the evaluation in simulation is performed for the 3D generalization of the KOP that addresses a broader range of real-world scenarios where a UAV needs to visit locations at varying heights. 
Therefore, the Tsiligirides 2 instance dataset is extended to 3D for validation of the planner performance.
The height assignments in the z-axis of each target are such that there is a volumetric sparsity in the reward distribution. 
The bounds on the time-step are $t_{\text{min}} = \SI{0.1}{\second}$ and $t_{\text{max}} = \SI{0.9}{\second}$.
The value of the nominal time-step $T_{n} = \SI{0.3}{\second}$ is used to compute the length of the MPC prediction horizon $N_h = C_{\text{max}} / T_{n}$.
An example of the planned smooth, feasible 3D trajectories along with the distribution of the sampled points is shown in \cref{fig:tsili_3d}.

\subsection{Real-world Deployments in 3D Persistent Monitoring}\label{sec:results_persistent}

A series of hardware experiments were performed to emulate a continual monitoring task with targets at known locations in two setups.
The first scenario is for static targets with dynamic reward gains.
In the second scenario, both targets and reward gains are dynamic.

\begin{figure}[!htb]\centering
   \includegraphics[width=1\linewidth]{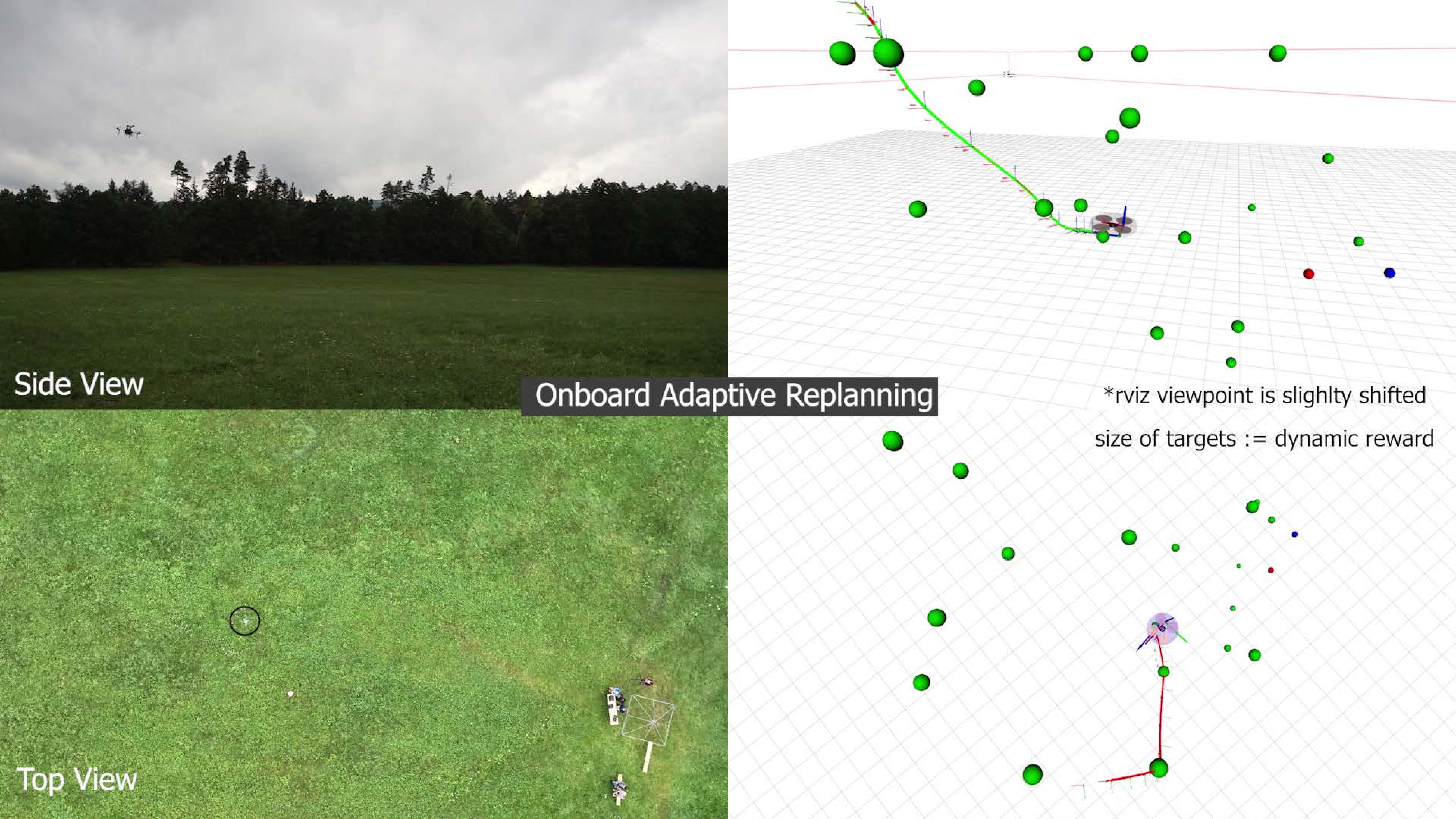}
   \caption{Experimental field deployment with static target position and dynamic target reward gains in a 3D persistent monitoring task with on-board re-planning.}
   \label{fig:real_world_expt}
\end{figure}
For the \textbf{static target with dynamic reward gains}, the value of the target reward gain increase factor is $\alpha_g=1$.
Synchronized views of the UAV during the experiment with the visualization of the targets in RViz~\cite{rviz15} are shown in ~\cref{fig:real_world_expt}.
The dynamic reward gains model is simulated onboard the UAV, which can be replaced by a perception module that would provide information about the latest target locations and particular reward gains.
The online re-planning is performed entirely onboard the UAV. 
Once collected, the target reward gain drops to 0 and increases as per~\cref{eq:reward_dynamics} such that targets with higher gains are prioritized and continual monitoring is achieved.

\begin{figure}[htb]
\vspace{-1em}
   \subfloat[The planned trajectory is not seen to intercept any moving targets at their current position.]{\includegraphics[width=0.31\columnwidth]{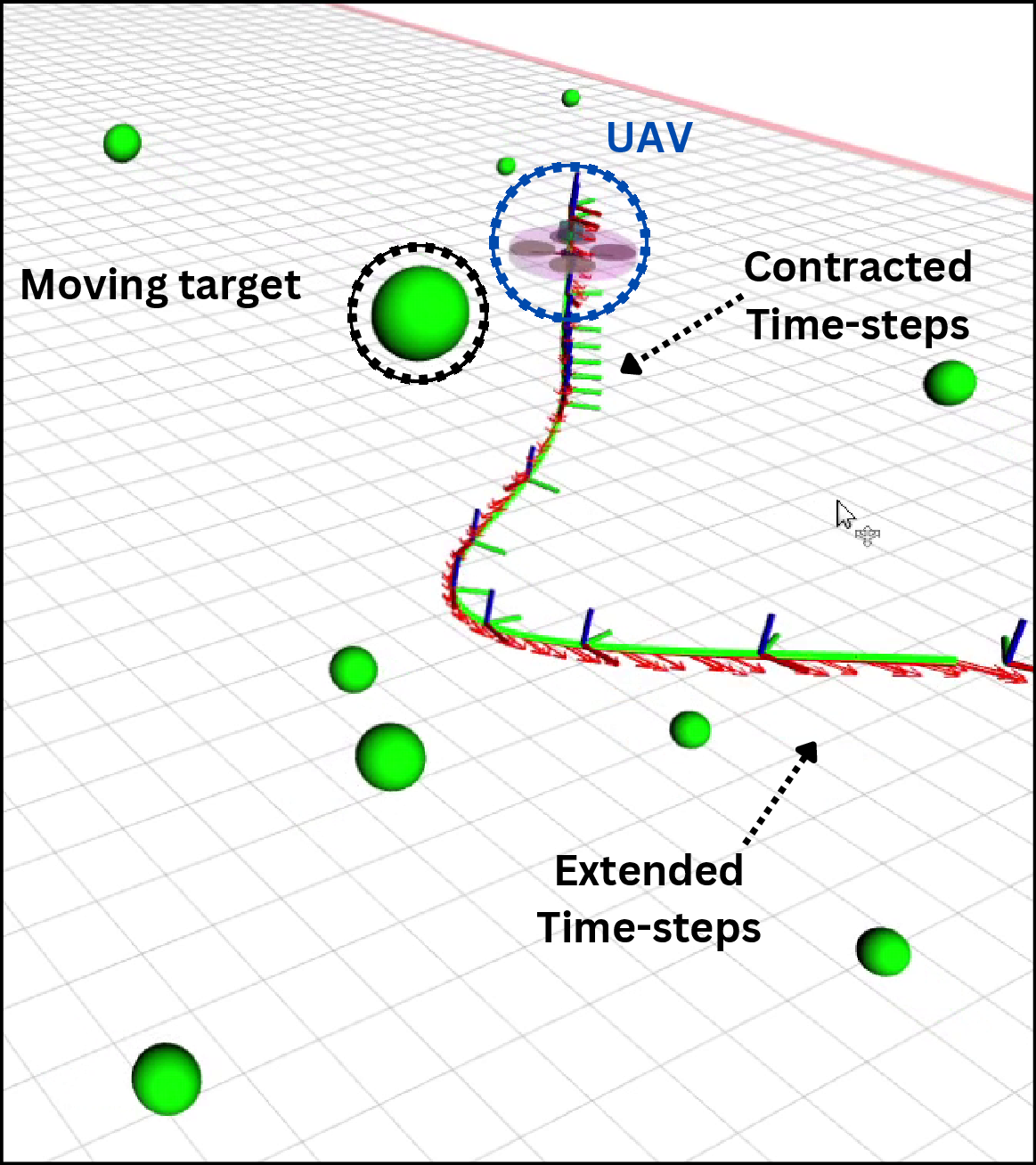}}
   \hfill
   \subfloat[The trajectory is planned with the predicted target position in consideration.]{\includegraphics[width=0.31\columnwidth]{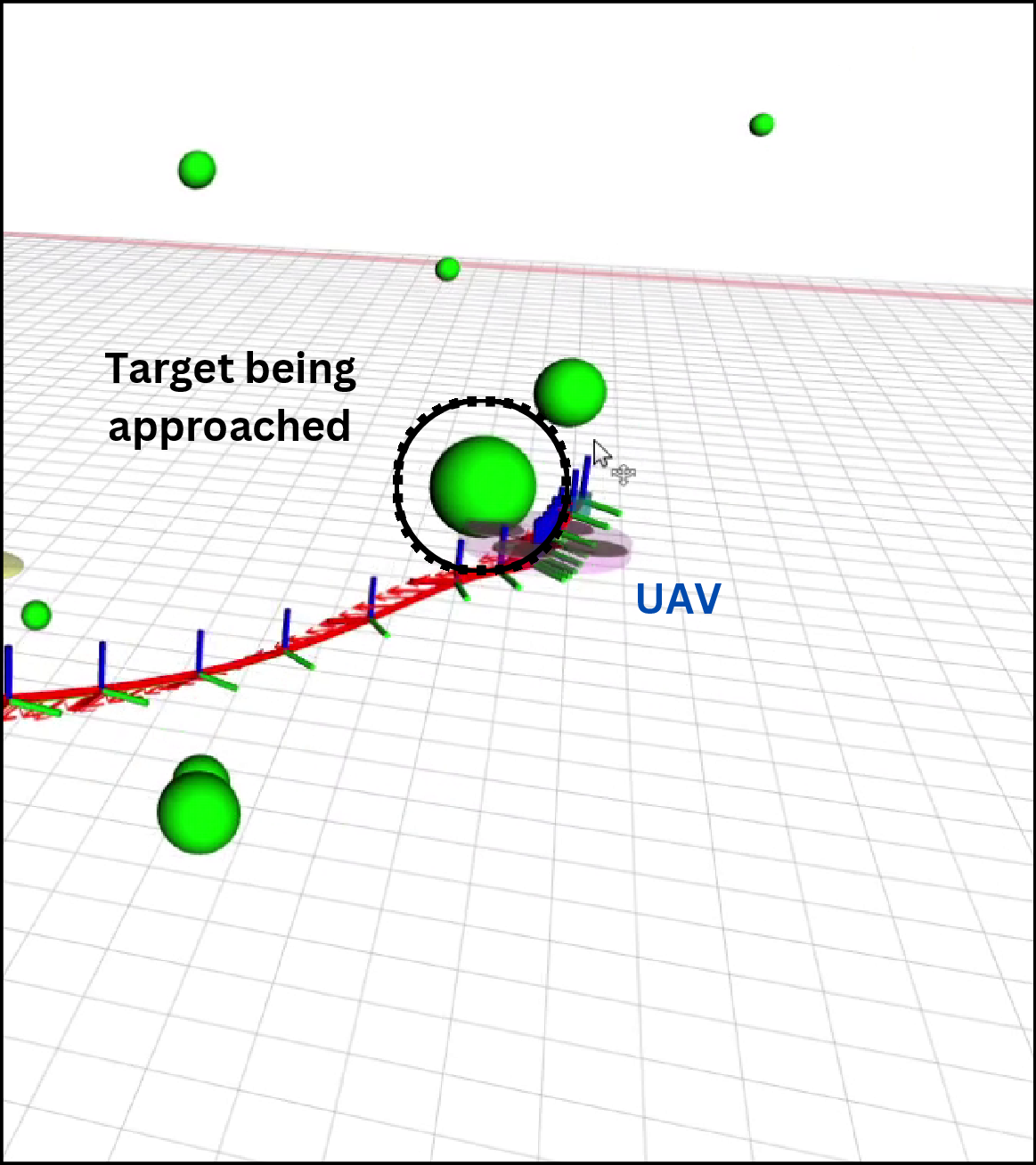}}
   \hfill
   \subfloat[The moving target is intercepted precisely, indicated by the reduced target sphere size.]{\includegraphics[width=0.31\columnwidth]{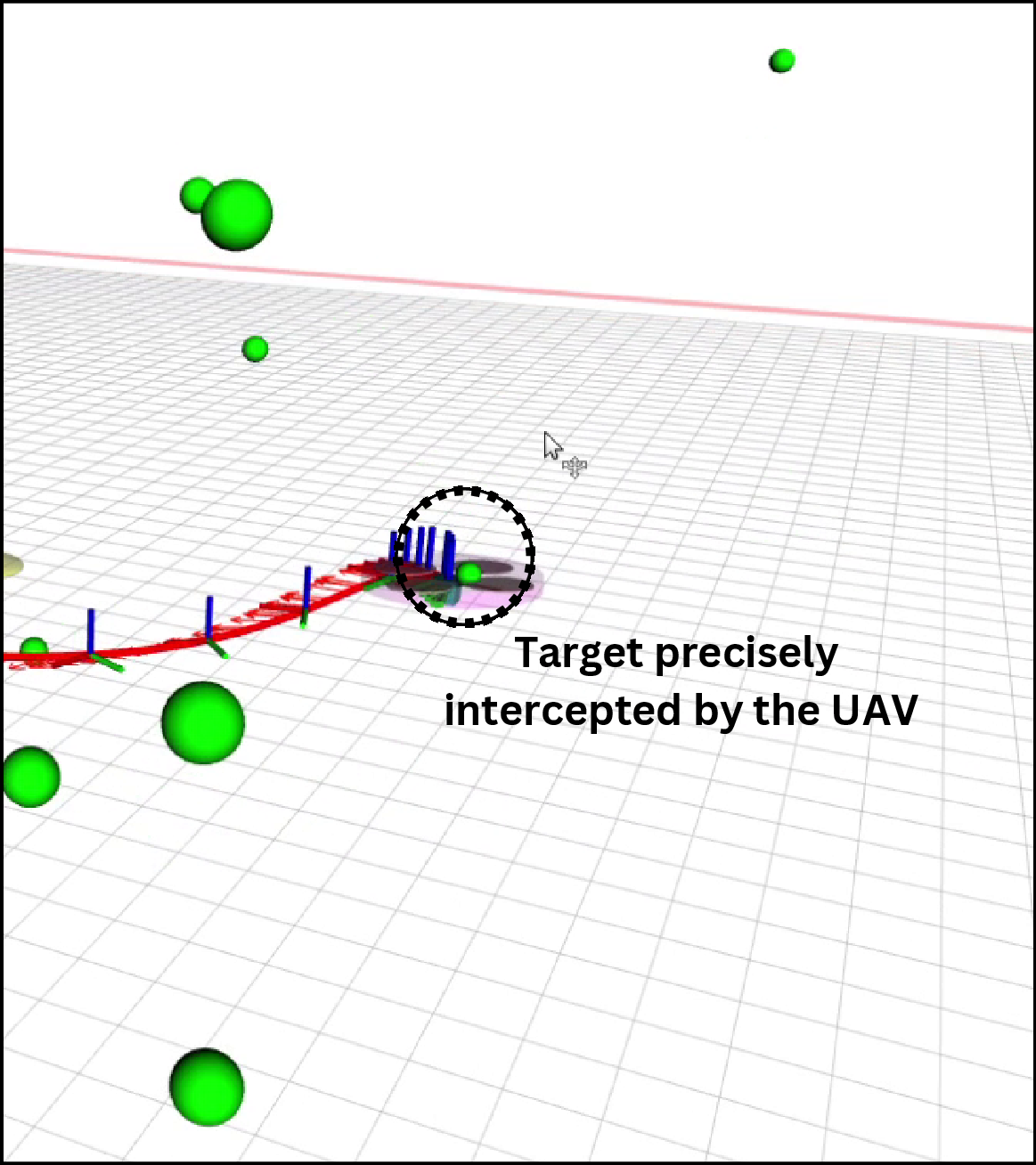}}
   \caption{Visualization of dynamic moving target interception by the vehicle in the 3D persistent monitoring task with on-board re-planning.}
   \label{fig:moving_targets_interception}
\end{figure}

The \textbf{dynamic target and reward gains} setups are designed to demonstrate the real vehicles' ability of precise moving target interception as shown in~\cref{fig:moving_targets_interception}.
The approaches are validated with parameters $N_h = 12$, $T_{h} = \SI{4.0}{\second}$, and $t_\text{replan}=\SI{0.9}{\second}$ for 3D, and $t_\text{replan}=\SI{0.5}{\second}$ for 2D.
The performance of the real-world flights is captured in the accompanying video available at \cite{vt-mpc}.

In 3D interception, simulated targets move around their initial positions following a time-parameterized trefoil knot motion as
\begin{equation}\label{eq:treefoil}
   \begin{array}{ll}
      x_r(t) &= \sin(t + t_r) + 2 \sin(2(t+t_r),\\
      y_r(t) &= \cos(t + t_r) - 2 \cos(2(t + t_r)),\\
      z_r(t) &= -3\sin(t + t_r).
   \end{array}
\end{equation}
The positions $x_r(t), y_r(t),z_r(t)$ are displacements of the $r$th target along the three axes from their initial positions.
A~phase difference $t_r$ is introduced to each target to ensure that the relative motion around their initial positions is not synchronized.
The value of $t_r$ for each of $N_t$ targets is calculated according to $t_r = \frac{2 \pi r}{n_t}$.

\begin{figure}[ht]\centering
   \vspace{-0.5em}
   \includegraphics[width=\linewidth]{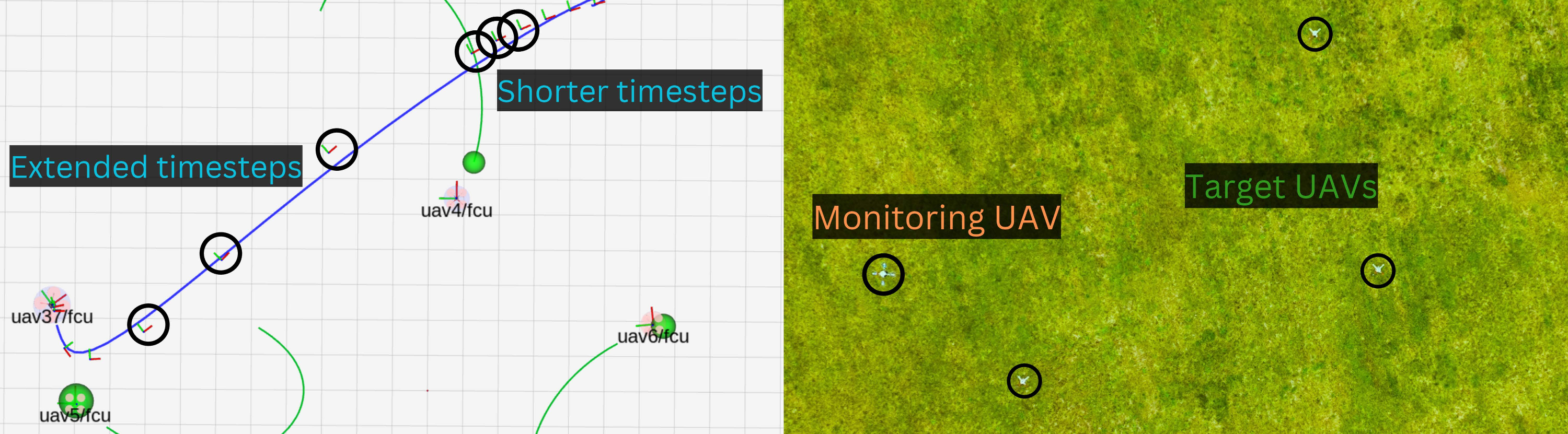}
   \caption{Experimental field deployment for 2D interception with real dynamic targets (target UAVs). The figure shows shorter time-steps (corresponding to precise maneuvers) planned in regions where target interception is predicted.}
   \label{fig:2d_monitor}
   \vspace{-1em}
\end{figure}

In 2D interception in the X-Y plane, three targets are represented by UAVs following plane-limited treefoil knot trajectories at a height of $\SI{2}{m}$ above the ground while the monitoring UAV is operated at $\SI{10}{m}$. ~\cref{fig:2d_monitor} shows a snapshot of the experimental setup.
Monitoring UAV velocity and acceleration constraints of $4$, $6$ and $\SI{8}{m/s}$, and $2$, $3$ and {$\SI{4}{\meter\per\second^2}$} respectively were used to demonstrate the principle of the VT-MPC formulation in real-world experiments.

Although a motion model for the targets is assumed to be known, the continual re-planning ability of the proposed approach allows the motion model to be estimated and updated online via an independent perception and target state estimation, which can run in parallel onboard the UAV.

\subsection{Dynamic Feasibility of Trajectories}\label{sec:results_dynamic}

\begin{figure}[ht]\centering
    \vspace{-0.5em}
    \includegraphics[width=\linewidth]{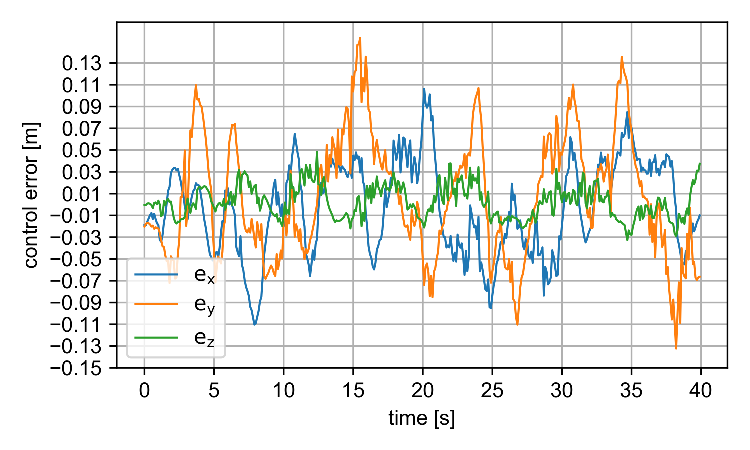}
    \caption{Control errors corresponding to the 3D Trajectory in~\cref{fig:cover-page}.}
    \label{fig:control_errors}
    \vspace{-1em}
\end{figure}

The feasibility of the planned trajectories produced by the proposed planner is verified to be dynamically feasible in multiple simulation instances and real-world experiments.
The trajectories are tracked by the MPC Tracker and SE(3) controller of the MRS UAV System~\cite{mrs_uav_sys}.
A sample tracking error profile is shown in~\cref{fig:control_errors} with the RMSE tracking errors of~\SI{0.0420}{\meter},~\SI{0.0557}{\meter}, and \SI{0.0149}{\meter} in $x$, $y$, and $z$ axes, respectively.
The performance of the error is similar in all deployments, and the error values are considered sufficiently small, supporting the feasibility of the planned trajectories.

\section{Conclusion}

In the paper, we present a novel MCP formulation for precise dynamic target interception utilizing variable time steps.
The reported results show that the proposed planner outperforms the state-of-the-art methods for the 2D Kinematic Orienteering Problem.
Furthermore, its extension to 3D instances enables adaptation to various tasks.
The proposed approach with the variable time steps exhibits two distinct behaviors.
\begin{enumerate}
    \item Short lengths of the time-step in areas with a high reward (information) gain for precise interception and maneuvering.
    \item Long lengths of the time-step in areas without rewards.
\end{enumerate}
A variable length of the time step enables a longer temporal length of the prediction horizon for the same number of steps than the fixed-length steps. 
Thus, it supports increasing the solution quality without exponentially increasing computational requirements caused by the increase in the number of prediction steps.
Based on the empirical evaluation using instances with static targets but with dynamic reward gains, the quality of solutions provided by the IPOPT solver has also increased with the variable time-step approach.
The results support that the proposed approach is viable for planning persistent monitoring and surveillance missions.
The reported results are backed by multiple real-world deployments that further support the practical feasibility of the proposed approach.
The possible future work includes generalization to multi-vehicle problems.

\bibliographystyle{IEEEtran}
\bibliography{main}
\end{document}